\documentclass[letterpaper, 10 pt, conference]{ieeeconf}  %

\IEEEoverridecommandlockouts
\usepackage{amsmath} %
\usepackage{dblfloatfix}
\usepackage[pdftex]{graphicx}
\graphicspath{{figures/}}
\usepackage{hyperref}
\hypersetup{
    colorlinks=true,
    linkcolor=black,
    citecolor=black,
    filecolor=black,
    urlcolor=black,
}
\usepackage[T1]{fontenc}
\usepackage[utf8]{inputenc}
\usepackage{csquotes}
\usepackage[english]{babel}
\usepackage[export]{adjustbox}
\usepackage{caption}
\usepackage{subcaption}
\usepackage{orcidlink}
\usepackage{hyperref}
\usepackage[capitalize]{cleveref}
\usepackage{nicefrac}
\usepackage{balance}
\usepackage{multirow}
\usepackage{booktabs}
\usepackage{makecell}
\usepackage{cuted}
\usepackage{siunitx}
\usepackage{xspace}
\usepackage[inkscapepath=out/svg-inkscape/,inkscapearea=page]{svg}
\usepackage[extractpath=out/svg-extract,extractname=filename]{svg-extract}
\svgpath{figures/}

\usepackage{placeins}%

\usepackage[style=ieee,
            doi=false,
            url=false,
            mincitenames=1,
            maxcitenames=1,
            minbibnames=6,
            maxbibnames=6,
            backend=biber]{biblatex}  %
\newcommand{\eg}{e.\,g.,\xspace}    %
\newcommand{\ie}{i.\,e.,\xspace}    %
\newcommand{\resp}{resp.\xspace}    %

\title{
Mosaic: An Extensible Framework for Composing\\
Rule-Based and Learned Motion Planners
}

\author{
    Nick Le Large\,\orcidlink{0009-0006-5191-9043}*$^{1}$,
    Marlon Steiner\,\orcidlink{0009-0005-4025-9142}*$^{1}$,
    Lingguang Wang\,\orcidlink{0000-0003-3833-9803}$^{1}$,
    Willi Poh\,\orcidlink{0009-0007-8217-6438}$^{1}$, \\
    Jan-Hendrik Pauls\,\orcidlink{0000-0003-2048-392X}$^{1}$,
    {\"{O}}mer {\c{S}}ahin Ta\c{s}\,\orcidlink{0000-0002-1249-260X}$^{2}$ and
    Christoph Stiller\,\orcidlink{0000-0003-4165-2075}$^{1}$%
    \thanks{* Authors have equal contribution.}
    \thanks{
       $^{1}$Institute of Measurement and Control Systems, Karlsruhe Institute of Technology (KIT),
       Karlsruhe, Germany
       {\tt\small \{firstname.lastname\}@kit.edu}
    }%
    \thanks{
       $^{2}$FZI Research Center for Information Technology, Karlsruhe, Germany
       {\tt\small \{lastname\}@fzi.de}
    }%
}

\usepackage{textcomp}
\usepackage[absolute,overlay]{textpos}
\usepackage{setspace}
\newcommand*{\presentationfont}{\fontfamily{lmss}\selectfont}

\newcommand{\presentationinformation}[2]{
\TPshowboxesfalse 
\begin{textblock}{0.82}(0.02,0.01)
    \vspace{2mm}
    \noindent{\bfseries{\presentationfont{\footnotesize{#1 \\
                                                        DOI: \href{http://dx.doi.org/#2}{#2}}}}}
\end{textblock}
}

\TPMargin{5pt}

\newcommand{\copyrightstatement}[1]{
    \TPshowboxesfalse
    \begin{textblock}{0.825}(0.089,0.94)
        \setstretch{0.65}%
        \noindent
        \begin{minipage}{\linewidth}
            {\fontsize{6.5pt}{8pt}\selectfont
                \parbox{\dimexpr\linewidth-0mm}{%
                    \copyright{} #1 IEEE.
                    Personal use of this material is permitted.
                    Permission from IEEE must be obtained for all other uses, in any current or future media,
                    including reprinting/republishing this material for advertising or promotional purposes,
                    creating new collective works, for resale or redistribution to servers or lists,
                    or reuse of any copyrighted component of this work in other works.
                }
            }
            \vspace{-1mm} %
        \end{minipage}
    \end{textblock}
}

\newcommand{\MosaicValCLSNR}{95.56}
\newcommand{\MosaicValCLSR}{94.18}
\newcommand{\MosaicInterPlanCLSR}{54.10}
\newcommand{\AvoidCrashedCarsMosaicCLSR}{29.80}

\newcommand{\MosaicCollisions}{16}
\newcommand{\VerifCollisionReductionPercent}{60.0}
\newcommand{\ScenarioTypesOutperform}{8}
\newcommand{\ScenarioTypesTied}{2}
\newcommand{\ScenarioTypesUnderperform}{4}
\newcommand{\LargestScenarioTypeRegression}{2.4}
\newcommand{\ZeroScoreReductionVsBestSinglePercent}{26.1}
\newcommand{\ScenarioTypesWonOrTied}{10}

\newcommand{\InterPlanImprovementOverFlowDrive}{22.8}
\newcommand{\InterPlanImprovementOverPDMClosed}{26.8}

\newcommand{\RuntimeMosaicPerScenarioSec}{2.0}
\newcommand{\RuntimeFlowDrivePerScenarioSec}{1.7}
\newcommand{\RuntimePDMClosedPerScenarioSec}{1.4}

\newcommand{\SelectionFlowDrivePercent}{18.6}
\newcommand{\SelectionPDMClosedPercent}{28.8}
\newcommand{\SelectionEqualPercent}{52.6}

\newcommand{\VerifFailPDMPercent}{0.45}
\newcommand{\VerifFailFlowDrivePercent}{0.37}
\newcommand{\VerifFailAtLeastOnePercent}{0.70}
\newcommand{\VerifFailBothPercent}{0.12}
\newcommand{\TotalTimesteps}{166806}
\newcommand{\EmergencyStopCount}{197}

\newcommand{\EmergencyDeceleration}{8.0}

\newcommand{\WeightComfortable}{2}
\newcommand{\WeightProgress}{5}
\newcommand{\WeightTTC}{7}
\newcommand{\ProgressGateThreshold}{0.2}

\newcommand{\SensitivityCLSRMin}{93.85}
\newcommand{\SensitivityCLSRMax}{94.18}
\newcommand{\SensitivityCLSRSpread}{0.33}

\newcommand{\AblationWithoutVerifCLSR}{92.82}
\newcommand{\AblationWithoutVerifColl}{40}
\newcommand{\AblationWithoutVerifZero}{38}
\newcommand{\AblationFlowDriveOnlyCLSR}{93.37}
\newcommand{\AblationFlowDriveOnlyColl}{15}
\newcommand{\AblationFlowDriveOnlyZero}{23}
\newcommand{\AblationFlowDriveOnlyEB}{0.31}
\newcommand{\AblationPDMClosedOnlyCLSR}{92.37}
\newcommand{\AblationPDMClosedOnlyColl}{17}
\newcommand{\AblationPDMClosedOnlyZero}{28}
\newcommand{\AblationPDMClosedOnlyEB}{0.14}
\newcommand{\AblationMosaicZero}{17}

\newcommand{\RawFlowDriveCLSR}{92.96}
\newcommand{\RawFlowDriveColl}{25}
\newcommand{\RawFlowDriveZero}{32}
\newcommand{\RawFlowDriveEB}{0.29}
\newcommand{\RawPDMClosedCLSR}{92.18}
\newcommand{\RawPDMClosedColl}{25}
\newcommand{\RawPDMClosedZero}{35}
\newcommand{\RawPDMClosedEB}{0.16}

\begin{document}

\presentationinformation{Presented at the IEEE/RSJ International Conference on Intelligent Robots and Systems (IROS) 2026}{10.1109/IROS61738.2026.11776088}
\maketitle

\copyrightstatement{2026}

\thispagestyle{empty}
\pagestyle{empty}

\begin{abstract}
    Safe and explainable motion planning remains a central challenge in autonomous driving.
    While rule-based planners offer predictable and explainable behavior,
    they often fail to grasp the complexity and uncertainty of real-world traffic. 
    Conversely, learned planners exhibit strong adaptability but suffer from reduced transparency and occasional safety violations. 
    We introduce \textit{Mosaic},
    a framework for structured decision-making that integrates both paradigms through arbitration graphs.
    By decoupling trajectory verification and selection from the generation of trajectories by individual planners,
    every decision becomes transparent and traceable.
    This separation lets verification and trajectory selection contribute independently:
    centralized verification acts as a safety floor,
    reducing at-fault collisions from \RawPDMClosedColl{} for each standalone planner to \MosaicCollisions{}.
    In contrast, per-step trajectory selection acts as a performance ceiling,
    combining the complementary strengths of a rule-based and a learned planner.
    In experimental evaluation on nuPlan, \textit{Mosaic} achieves $\mathbf{\MosaicValCLSNR}$ CLS-NR
    and $\mathbf{\MosaicValCLSR}$ \mbox{CLS-R} on the Val14 closed-loop benchmark, setting a new state of the art.
    On the interPlan benchmark, focused on highly interactive and out-of-distribution scenarios,
    \textit{Mosaic} scores $\mathbf{\MosaicInterPlanCLSR}$ \mbox{CLS-R},
    outperforming its best constituent planner by \InterPlanImprovementOverFlowDrive\%---all without retraining or requiring additional data.
    The code is available at \href{https://github.com/KIT-MRT/mosaic}{\nolinkurl{github.com/KIT-MRT/mosaic}}.
\end{abstract}
\section{Introduction}
\label{introduction}

Motion planning is a fundamental task for autonomous driving, generating safe, comfortable, and goal-directed trajectories in dynamic traffic.
Achieving these objectives simultaneously remains challenging, as planners must adapt to diverse scenarios while remaining explainable to ensure trust and verifiability.

Existing approaches fall into three main categories. Rule-based planners \cite{treiber2000idm, dauner2023pdm} encode hand-designed heuristics to produce predictable, explainable behavior, but struggle to adapt to complex, unstructured, or unseen situations. Learning-based planners \cite{dauner2023pdm, cheng2024pluto, liao2024diffusiondrive, wang2025flowdrivemoderatedflowmatching, cusumanotowner2025robustautonomyemergesselfplay}, in contrast, adapt well by leveraging data-driven models to capture the variability of real-world driving.
However, their decision-making is not directly accessible, making outputs hard to interpret or verify---a limitation for safety-critical deployment.
Hybrid approaches \cite{dauner2023pdm, huang2023gameformer, cheng2024pluto} typically combine learned planners with rules as post-processing, but are mostly non-unified and tailored to a specific use case.

\begin{figure}[ht!]
    \centering
    \includegraphics[width=0.8\linewidth]{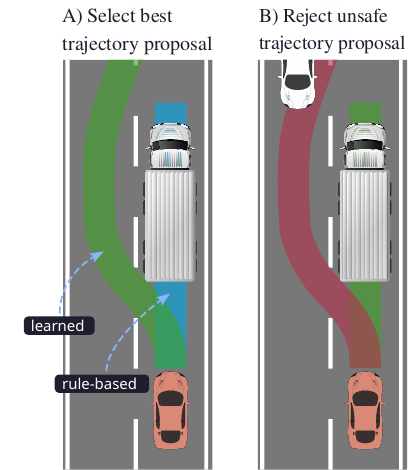}
    \caption{\textit{Mosaic} combines complementary planners---\eg rule-based and learned---within a structured and extensible framework. An arbitrator selects the best trajectory (\textit{green}) from candidate proposals (\textit{blue}) while a shared verification step rejects unsafe plans (\textit{red}), enabling safe and explainable motion planning.}
    \label{fig:motivation}
\end{figure}

In this work, we propose a unified framework for motion planning based on arbitration graphs (AGs) \cite{lauer2010cognitive, spieker2025arbitrationgraphs}, which are a structured decision-making mechanism that enables coordination among multiple behavior components.
We call our framework \textit{Mosaic} (see \cref{fig:motivation}), as it combines heterogeneous planning approaches.
Much like the art form that assembles diverse pieces into a beautiful, coherent whole that is greater than the sum of its parts.

While many state-of-the-art motion planners include some kind of post-processing or internal verification, we propose to structurally separate trajectory generation, the strengths of individual planning approaches, from trajectory verification and selection.
Building upon the arbitration concept, a cost arbitrator can evaluate the trajectory proposals of two or more complementary planners, \eg a rule-based and a learned approach.
Within our work, we wrap the learning-based planner FlowDrive*~\cite{wang2025flowdrivemoderatedflowmatching} (the * refers to the hybrid version of the planner) and the rule-based planner PDM-Closed~\cite{dauner2023pdm} into the AG.

Through unified verification, unsafe proposals that would lead to collisions can be rejected,
while a fallback behavior (emergency braking) integrates naturally as a safety layer within the arbitration graph.
A unified scoring function, inspired by the nuPlan metric, allows choosing the best proposal and thus combining the strengths of both approaches.

Finally, the same interface is designed to extend to additional motion planners and fallback layers.
At the same time, it makes it clear which trajectory is chosen and why, \ie due to cost or validity,
enabling not only safe and high-performing but also explainable decision-making for autonomous driving.

Our contributions are as follows:
\begin{itemize}
    \item We present \textit{Mosaic},
    a motion planning framework that decouples trajectory \emph{verification} and \emph{selection} from trajectory \emph{generation}
    using arbitration graphs.
    Any planner---rule-based, learned, or hybrid---is wrapped as a black-box behavior component behind a common interface,
    so the decision of how much to trust each planner becomes a transparent,
    per-step runtime judgment rather than logic hard-coded inside a bespoke pipeline.
    \item We show that verification and selection contribute independently.
    Centralized verification is the safety mechanism:
    it reduces at-fault collisions from \RawPDMClosedColl{} for each standalone planner to \MosaicCollisions{}.
    Per-step trajectory selection is the performance mechanism:
    it lifts closed-loop driving performance above either planner alone,
    reduces zero-score scenarios by at least \SI{\ZeroScoreReductionVsBestSinglePercent}{\percent} over either planner alone,
    and matches or outperforms both planners in \ScenarioTypesWonOrTied{} of 14 scenario types.
    \item We achieve state-of-the-art performance on the nuPlan closed-loop benchmarks
    (Val14 split~\cite{dauner2023pdm}) with $\MosaicValCLSNR$ CLS-NR and $\MosaicValCLSR$ CLS-R.
    On the out-of-distribution interPlan benchmark, we reach $\MosaicInterPlanCLSR$ CLS-R,
    a \SI{\InterPlanImprovementOverFlowDrive}{\percent} gain over the best constituent planner---%
    all without retraining the underlying planners or requiring additional data.
\end{itemize}

\section{Related Work}
\label{sec:related_work}

Autonomous driving requires reliable motion planning capable of handling diverse environments, uncertain interactions, and safety-critical decisions.  
This section reviews prior work on motion planning paradigms and on decision-making frameworks based on AGs, highlighting the need for a unified and interpretable integration mechanism across potentially heterogeneous motion planners.

\subsection{Motion Planning}

Motion planning for autonomous driving can be broadly categorized into \textit{rule-based}, \textit{learning-based}, and \textit{hybrid} approaches, with recent trends extending toward generative and language-based models.  

\textit{Rule-based methods} rely on deterministic algorithms~\cite{treiber2000idm}, handcrafted rules~\cite{dauner2023pdm}, or optimization-based formulations~\cite{tas2023decisiontheoretic} to generate feasible and interpretable trajectories.   
While such methods offer transparency and predictability, they typically struggle in highly dynamic or unstructured environments.

\textit{Learning-based approaches}~\cite{cheng2023plantf, cheng2024pluto, zhou2024behaviorgpt, wu2024smart} leverage neural architectures, often transformers, to learn motion planning directly from data.  
These models exhibit adaptability and generalization across complex scenarios.  
In addition to imitation learning, reinforcement learning (RL) \cite{konstantinidis2024AIL, cusumanotowner2025robustautonomyemergesselfplay}, where agents learn driving policies through interaction and reward feedback, has been widely explored for motion planning. 
RL-based planners have shown promise in closed-loop control \cite{cusumanotowner2025robustautonomyemergesselfplay} and adaptation to rare scenarios, but often require extensive training and safety constraints.
End-to-end approaches~\cite{li2025navigationguidedsparsescenerepresentation, yuan2024mamba, chen2024drivinggpt, nishimura2023rap} further integrate perception into the planning task, directly mapping sensor inputs to control commands or trajectories. 
Recently, generative models such as diffusion- or flow-based planners~\cite{wang2025flowdrivemoderatedflowmatching, huang2025gen, hu2024mpwithgenmodel, liao2024diffusiondrive, zheng2025diffusionbased} have emerged as powerful tools for producing diverse, realistic, and multimodal trajectory distributions.  
In parallel, language-driven methods explore the integration of natural language reasoning into driving systems, leveraging large language models (LLMs) ~\cite{chen2024asynchronous, li2024driving} and vision language models (VLMs)~\cite{sima2024drivelm, pan2024vlp} to explain semantic instructions and contextual cues.  

Despite their variety, most learning-based planning paradigms rely on \textit{hybrid methods} \cite{huang2023gameformer, zheng2025diffusionbased, dauner2023pdm} with a fixed internal decision process as post-processing to choose or refine the final trajectory.
Note that there are also approaches utilizing learned methods to refine the rule-based trajectory~\cite{dauner2023pdm, fischer2023srmpc}.
When designing such a hybrid method, the decision logic is often ad hoc, non-modular, or hard to reproduce. 
This lack of a unified arbitration mechanism limits interpretability, extensibility, and safe integration of diverse planning modules.

\subsection{Decision-Making and Arbitration Graphs}

\begin{figure*}[!t]
    \vspace*{2mm}
    \centering
    \includegraphics[width=\linewidth]{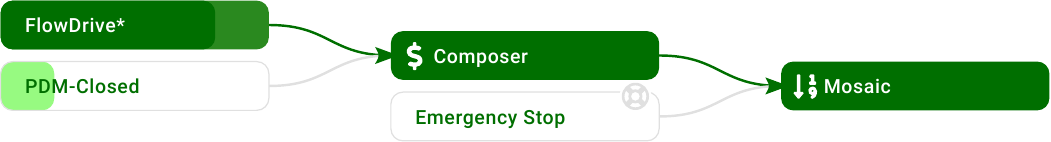}
    \caption{The proposed AG structure with 3 behavior components (\texttt{FlowDrive*}, \texttt{PDM-Closed}, \texttt{Emergency Stop}), a cost arbitrator \texttt{Composer}, and a priority arbitrator \texttt{Mosaic}.}
    \label{fig:AG-structure}
\end{figure*}

AGs provide a structured and explainable mechanism for hierarchical decision-making in autonomous systems \cite{spieker2025arbitrationgraphs}.  
Originally developed in the context of robotic soccer~\cite{lauer2010cognitive} and later used in autonomous driving \cite{orzechowski2020decision}, AGs draw inspiration from Brooks’ behavior-based subsumption, knowledge-based architectures like Belief-Desire-Intention, and object-oriented design principles \cite{spieker2025arbitrationgraphs}. 

In this framework, atomic \textit{behavior components} represent fundamental abilities or behaviors that interpret the current situation and propose \textit{commands} when their \textit{invocation conditions} are met.
Arbitrators organize these behaviors hierarchically, filtering applicable options and selecting the most suitable one for execution using strategies such as priority- or cost-based selection.  
Compared to Behavior Trees, AGs explicitly encode preconditions and decision logic, leading to greater robustness and transparency.

Recent developments extend AGs with domain-specific \textit{verification mechanisms} that validate the safety and feasibility of proposed commands before execution \cite{spieker2025arbitrationgraphs}.  
This enables the integration of experimental or learned behaviors without compromising system reliability, as unsafe or invalid outputs are intercepted in a verification step and mitigated through structured fallback layers.

However, prior work has not explored the use of arbitration graphs on real-world data to evaluate the effectiveness of this approach on a widely recognized benchmark.
In contrast, our work applies AGs to the large-scale autonomous driving benchmark nuPlan \cite{caesar2021nuplan} and demonstrates that they can serve as an explainable decision structure capable of integrating heterogeneous planners within a single framework.

Notably, AGs can serve not only as a method to combine multiple planners, but also to wrap a single planner, \eg a learning-based one, into an explainable framework.
Therefore, AGs are also suitable for replacing opaque, non-unified and specifically tailored decision-making, which is built on top of most of the current learning-based planning approaches \cite{dauner2023pdm, cheng2024pluto, zheng2025diffusionbased, huang2023gameformer}.

\section{The Mosaic Framework}
\label{sec:methodology}

The overall architecture of \textit{Mosaic} is shown in \cref{fig:AG-structure}. 
We make use of the explainable AG concept to compose two planners into a single architecture.
Briefly, the framework can be understood as follows (from left to right, see \cref{fig:AG-structure}):

\begin{itemize}
    \item \texttt{FlowDrive*} / \texttt{PDM-Closed}: These components correspond to behaviors in the AG framework, each proposing a trajectory given the current state of the world (see \cref{subsec:methodolgy/behaviors}).
    \item \texttt{Composer}: This element is a cost arbitrator (see \cref{subsec:methodolgy/costarbitrator}) with an integrated verification step. For each behavior, the arbitrator verifies the proposed trajectory (see \cref{subsec:methodolgy/verification}); only verified trajectories are scored and considered for selection.
    \item \texttt{Emergency Stop}: This behavior computes an emergency braking trajectory. As indicated by the safety buoy, this is a last resort fallback which does not need to pass verification~\cite{spieker2025arbitrationgraphs}.
    \item \texttt{Mosaic}: This is the root arbitrator and it is designed as a priority arbitrator (see \cref{subsec:methodolgy/prioarbitrator}). If the \texttt{Composer} arbitrator provides a valid trajectory, this trajectory is chosen, otherwise the \texttt{Emergency Stop} trajectory is applied.
\end{itemize}

Throughout this paper, we use \texttt{monospace} font to refer to behavior components within the AG, which wrap the respective planner and include the modifications described in \cref{subsec:methodolgy/verification}.
Plain text refers to the underlying planner itself as presented in the original publications.

\subsection{Trajectory Planners as Behaviors in the AG}
\label{subsec:methodolgy/behaviors}
As \textit{Mosaic} is planner-agnostic, we demonstrate it with two complementary planners in the AG.
To combine the strengths of different motion planning paradigms, we choose a rule-based planner (PDM-Closed) and a learning-based planner (FlowDrive*).
While FlowDrive~\cite{wang2025flowdrivemoderatedflowmatching} is the current open-source SOTA model, PDM-Closed achieves strong results as well on the nuPlan closed-loop benchmark.
A rule-based and learning-based motion planner combination was already introduced through PDM-Hybrid (PDM-Closed + PDM-Open) \cite{dauner2023pdm}, achieving slightly worse results in the closed-loop benchmark than PDM-Closed as a stand-alone. 
Contrary to this, we propose the use of a generic AG framework to fuse different planners into an explainable structure, naturally including trajectory verification and scoring, as well as fallback maneuvers.
In the following, we provide a brief description of the utilized planners.

\subsubsection{Learning-based FlowDrive Planner}
The FlowDrive model \cite{wang2025flowdrivemoderatedflowmatching} utilizes rectified flow-matching \cite{lipman_flow_2023, liu_flow_2022} for trajectory generation. This model addresses the common issue of data imbalance in driving datasets.
While FlowDrive represents a purely learned approach based on a diffusion transformer architecture \cite{peebles_scalable_2023}, the authors also provide an extension of the model using post-processing, resulting in a hybrid planning method (FlowDrive*).
FlowDrive* features a moderated guidance mechanism that injects small perturbations during trajectory generation to increase maneuver diversity, such as overtaking or nudging. 
Evaluated on the nuPlan benchmark, FlowDrive* achieves SOTA results among open-source methods, surpassing previous learning-based and hybrid methods.
We utilize FlowDrive* in a behavior component of the AG presented in this work.

\subsubsection{Rule-based PDM-Closed Planner}
The Predictive Driver Model Closed (PDM-Closed) \cite{dauner2023pdm} is a rule-based planner that extends the established Intelligent Driver Model (IDM). 
It is designed to maximize performance in the nuPlan closed-loop evaluation task.
The model incorporates concepts from model predictive control, involving forecasting, proposal generation, simulation, scoring, and trajectory selection. 
It first uses a graph search to select a sequence of lane centerlines to follow.
To handle diverse scenarios and address the trade-offs found when selecting a single target speed, PDM-Closed generates 15 trajectory proposals. These proposals are created by implementing IDM policies across five distinct target speeds (20\% to 100\% of the speed limit) at three lateral centerline offsets (±1 m and 0 m). The environment forecasting uses a simple constant velocity forecast for other dynamic agents.
The trajectory proposals are simulated based on the expected closed-loop movement (using an LQR controller and a kinematic bicycle model). 
They are then scored based on metrics favoring traffic-rule compliance, progress, and comfort. 
An emergency brake is enforced if a collision is anticipated within two seconds.
At publication time, PDM-Closed achieved SOTA closed-loop performance on nuPlan, while its open-loop performance is very poor.

\subsection{Behavior Verification}
\label{subsec:methodolgy/verification}
\begin{figure}[htbp]
    \centering
    \includegraphics[width=\linewidth]{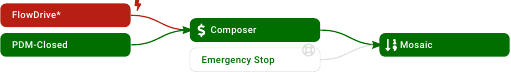}
    \caption{Behavior verification: One behavior component fails verification (\textit{red}) and is rejected. The remaining behavior is selected.}
    \label{fig:arbitration_verification}
\end{figure}

Both PDM-Closed and FlowDrive* include the same internal safety check:
if an at-fault collision is predicted within \SI{2.0}{\second} \emph{and} the ego speed is below \SI{5}{\metre\per\second},
the planner overrides its output with an emergency brake at \SI{4.05}{\metre\per\square\second}.
In \textit{Mosaic}, we disable these per-planner checks and replace them with a single verifier inside the \texttt{Composer}.
As introduced in~\cite{spieker2025arbitrationgraphs}, verification is integrated directly into the arbitration algorithm:
before an option is selected, the arbitrator performs a domain-specific verification of its proposed command.
If the command fails, it is discarded and the arbitrator proceeds to the next best option (see \cref{fig:arbitration_verification}).
Our verifier applies the same at-fault collision check to every candidate,
but \emph{without the speed gate}, so it fires uniformly at all velocities.
Agents are propagated with the same constant-velocity forecast that PDM-Closed uses internally.

Centralizing the check has two consequences.
Architecturally, components no longer carry their own safety logic,
so any planner can be integrated as long as the shared verifier's assumptions suffice,
and every rejection is attributed to a specific component and reason
(the fallback when both fail is described in \cref{subsec:methodolgy/prioarbitrator}).
Functionally, removing the speed gate and using the stronger emergency deceleration introduced in \cref{subsec:methodolgy/prioarbitrator}
makes the shared verifier \emph{stricter} than the original internal checks.
We isolate the safety impact of that change in \cref{subsec:ablations}.
We deliberately keep the verifier simple for this benchmark study.
Strengthening it toward a production safety module fits the same interface without touching the arbitration logic.

\subsection{The Cost Arbitrator}
\label{subsec:methodolgy/costarbitrator}

After filtering unsafe trajectories, the cost arbitrator (see \texttt{Composer} in \cref{fig:AG-structure}) applies a cost function to select the best remaining option.
Our scoring function is inspired by the nuPlan closed-loop evaluation metrics but, crucially, replaces their binary pass/fail decisions with \textit{continuous} scores in $[0,1]$, giving the cost arbitrator fine-grained information to distinguish between candidate trajectories.
Of course, unlike the nuPlan evaluation metrics, our scores have to be computed \textit{online} using only historical observations and predictions and cannot rely on ground truth data.
We calculate a total score $S_\mathrm{total}$ for each trajectory using a multiplicative gating approach:
\begin{equation}
S_{\mathrm{total}} = \underbrace{S_\mathrm{coll} \cdot S_\mathrm{driv} \cdot S_\mathrm{dir}}_{\text{safety gate}} \cdot\; G_\mathrm{prog} \cdot S_\mathrm{perf}.
\label{eq:total_score}
\end{equation}
The cost is $C_\mathrm{total} = -S_\mathrm{total}$.
All components are in $[0,1]$.

\textbf{Safety gate.}
The three safety scores act as a multiplicative gate:
if any is zero, then $S_\mathrm{total}=0$, so an unsafe trajectory is never preferred regardless of its performance score.
The collision score penalizes at-fault collisions by the normalized overlap between
the ego box $B_\mathrm{ego}(t)$ and each colliding object $B_k(t)$,
taking the worst case over the simulated time steps $t$ and objects $k$:
\begin{equation}
S_\mathrm{coll} = \min_{t,k}\left(1 - \frac{|B_\mathrm{ego}(t)\cap B_k(t)|}{|B_\mathrm{ego}(t)|}\right),
\end{equation}
so a grazing contact scores near $1$ and a full overlap $0$ (floored at $0.5$ for static obstacles).
$S_\mathrm{driv}$ is the fraction of the simulated poses inside the drivable area,
and $S_\mathrm{dir}$ applies discrete thresholds ($1.0/0.5/0.0$) to the longest contiguous distance driven against traffic.

\textbf{Progress gate.}
A soft gate $G_\mathrm{prog} = \min(d_\mathrm{route}/\tau,\; 1)$ penalizes standstill trajectories,
where $d_\mathrm{route}$ is the ego's displacement along the route centerline normalized by the expected distance at the current speed.

\textbf{Performance score.}
The performance score is a weighted average $S_\mathrm{perf}=\left(\sum_k w_k S_k\right)/\sum_k w_k$
over $k\in\{\mathrm{prog},\mathrm{ttc},\mathrm{comf}\}$, with the weights in \cref{tab:parameters}.
Time-to-collision (TTC) receives the highest weight as the most immediate safety signal available at scoring time,
progress captures the primary planning objective, and comfort acts as a secondary tiebreaker.
The TTC sub-score is the earliest predicted collision time under a constant-velocity ego rollout,
clamped to a \SI{3}{\second} horizon and normalized to $[0, 1]$.
Comfort checks acceleration and jerk limits.
Crucially, the verifier and this scorer are \emph{non-redundant}:
the verifier applies a binary at-fault check over a \SI{2}{\second} horizon,
whereas the scoring safety gate adds finer continuous penalties over the full horizon
and covers drivable-area and wrong-way violations that the verifier ignores.

\begin{table}[t]
    \vspace*{2mm}
    \caption{Scoring function weights within the cost arbitrator.}
    \centering
    \normalsize
    \begin{tabular}{l|r}
         Parameter & Value \\
         \hline
         $w_\mathrm{comfortable}$ & \WeightComfortable \\
         $w_\mathrm{progress}$ & \WeightProgress \\
         $w_\mathrm{ttc}$ & \WeightTTC \\
         Progress gate threshold $\tau$ & \ProgressGateThreshold \\
    \end{tabular}
    \label{tab:parameters}
\end{table}

\begin{table*}[ht!]
    \vspace*{2mm}
    \centering
    \caption{Comparison on the nuPlan Val14 and interPlan evaluation splits. CLS-NR and CLS-R denote the score for the closed-loop non-reactive and closed-loop reactive benchmark, respectively. Best results per evaluation type and per column are in \textbf{bold}, second-best are \underline{underlined}. $^\dagger$Closed-source model.}
    \label{tab:nuplan_results}
    \normalsize
    \begin{tabular}{llcc|c}
        \toprule
        \multirow{2}{*}{\textbf{Type}} & \multirow{2}{*}{\textbf{Planner}}
        & \multicolumn{2}{c}{\textbf{Val14}} & \textbf{interPlan}\\
         &  & CLS-NR $\uparrow$ & CLS-R $\uparrow$ & CLS-R $\uparrow$ \\
        \midrule
        Expert & Log-replay & 93.53 & 80.32 & 14.76 \\
        \midrule
        \multirow{4}{*}{Learning-based}
        & PDM-Open \cite{dauner2023pdm} & 53.53 & 54.24 & 25.02 \\
        & Diffusion Planner \cite{zheng2025diffusionbased} & 89.87 & 82.38 & 24.71  \\
        & FlowDrive \cite{wang2025flowdrivemoderatedflowmatching} & 91.21 & 85.37 & 36.96 \\
        & GIGAFLOW$^\dagger$ \cite{cusumanotowner2025robustautonomyemergesselfplay} & - & \underline{93.8} & - \\
        \midrule
        \multirow{6}{*}{Rule-based \& Hybrid}
        & IDM \cite{treiber2000idm} & 75.60 & 77.33 & 31.20 \\
        & PDM-Closed \cite{dauner2023pdm} & 92.84 & 92.12 & 41.23 \\
        & PDM-Hybrid \cite{dauner2023pdm} & 92.77 & 92.11 & 41.61 \\
        & Diffusion Planner w/ refine. \cite{zheng2025diffusionbased}& 94.26 & 92.90 & - \\
        & FlowDrive* \cite{wang2025flowdrivemoderatedflowmatching}& \underline{94.81} & 92.96 & \underline{44.05} \\
        & Mosaic (ours)& \textbf{\MosaicValCLSNR} & \textbf{\MosaicValCLSR} & \textbf{\MosaicInterPlanCLSR} \\
        \bottomrule
    \end{tabular}
\end{table*}
 
\subsection{Priority Arbitrator to Integrate Fallback Layer}
\label{subsec:methodolgy/prioarbitrator}

\begin{figure}[!h]
    \centering
    \includegraphics[width=\linewidth]{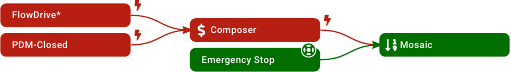}
    \caption{Emergency fallback: Both behavior components fail verification. The \texttt{Composer} becomes inapplicable, and the priority arbitrator falls back to \texttt{Emergency Stop}.}
    \label{fig:arbitration_emergency}
\end{figure}

As the last step in our AG design, we use a priority arbitrator (\texttt{Mosaic} in \cref{fig:AG-structure}), which selects valid options top-down.
A trajectory from the cost arbitrator is always preferred over the \texttt{Emergency Stop} behavior.
If neither \texttt{FlowDrive*} nor \texttt{PDM-Closed} passes verification, the \texttt{Composer} arbitrator becomes inapplicable and the priority arbitrator falls back to the \texttt{Emergency Stop} behavior (see \cref{fig:arbitration_emergency}).
Following~\cite{spieker2025arbitrationgraphs}, the emergency stop serves as the last resort fallback and does not need to pass verification.
It is implemented as a straight-line deceleration, ensuring a simple and deterministic action.
Compared to the emergency stop utilized in both FlowDrive* and PDM-Closed,
we increase the deceleration from \SI{4.05}{\metre\per\square\second} to \SI{\EmergencyDeceleration}{\metre\per\square\second}.
The original \SI{4.05}{\metre\per\square\second} coincides with nuPlan's \emph{comfort} deceleration threshold,
which is appropriate for normal driving but conservative for a last-resort maneuver.
Because this stronger brake is part of the shared verification stack,
we quantify its safety contribution together with the verifier in \cref{subsec:ablations}.

\section{Evaluation on nuPlan}
\label{sec:results_and_evaluation}

We evaluate our proposed \textit{Mosaic} framework on the nuPlan closed-loop benchmark and the interPlan benchmark, comparing against a wide range of motion planners, including the current learned SOTA method GIGAFLOW \cite{cusumanotowner2025robustautonomyemergesselfplay}.

\subsection{nuPlan Closed-loop Benchmark}
\label{sec:eval_closed_loop}
Following the evaluation established by \cite{dauner2023pdm}, we evaluate the Val14 split of the nuPlan benchmark, which comprises \num{1118} scenarios across 14 scenario types.
The closed-loop benchmark computes two composite scores:
CLS-NR (closed loop, non-reactive), where surrounding agents replay their recorded trajectories,
and CLS-R (closed loop, reactive), where they respond to the ego vehicle via an IDM policy~\cite{treiber2000idm}.
Each score lies in $[0, 100]$ and combines sub-scores for
at-fault collisions, drivable-area compliance, driving-direction compliance, ego progress along the route,
time-to-collision, speed-limit compliance, and comfort.
The safety-critical sub-scores act as multiplicative gates that drive the scenario score to zero on violation,
while the rest contribute through a weighted average.
The benchmark score is the mean closed loop score over all scenarios.
We additionally evaluate on the interPlan benchmark~\cite{hallgarten2024interplan},
a publicly available nuPlan extension focused on highly interactive and out-of-distribution scenarios.

\subsection{Baselines}
We compare against a comprehensive set of learned, rule-based, and hybrid planning methods.
In the category of purely learned planners, the current open-source SOTA is FlowDrive~\cite{wang2025flowdrivemoderatedflowmatching}.
The closed-source \mbox{GIGAFLOW}~\cite{cusumanotowner2025robustautonomyemergesselfplay} model is also a learned planner, trained purely in self-play simulation within a multi-agent reinforcement learning setting without seeing any data from the dataset.
In the category of hybrid methods, \mbox{FlowDrive*} with moderated guidance and a trajectory selection process leads the benchmark of open-source models.

\subsection{Quantitative Results}
\label{subsec:evaluation/results}

Table~\ref{tab:nuplan_results} summarizes the results on the Val14 benchmark. Among purely learning-based planners, diffusion-based \cite{zheng2025diffusionbased} and flow-matching \cite{wang2025flowdrivemoderatedflowmatching} models achieve the highest closed-loop scores, surpassing earlier methods \cite{cheng2023plantf, cheng2024pluto}. However, a consistent trend can be observed across all entries: when a rule-based component is integrated, forming a hybrid system, the resulting performance increases substantially \cite{cheng2024pluto, zheng2025diffusionbased, wang2025flowdrivemoderatedflowmatching}.
This highlights the crucial role of rule-based reasoning in enhancing the stability and safety of data-driven planners. Consequently, hybrid planning strategies currently represent the most effective class of motion planners on nuPlan.

Our proposed \textit{Mosaic} framework achieves $\MosaicValCLSNR$ CLS-NR and $\MosaicValCLSR$ CLS-R,
surpassing all baselines, including the closed-source GIGAFLOW,
setting a new SOTA on both the non-reactive and reactive closed-loop benchmarks.

On interPlan, \textit{Mosaic} achieves $\MosaicInterPlanCLSR$ \mbox{CLS-R},
outperforming FlowDrive* by $\SI{\InterPlanImprovementOverFlowDrive}{\percent}$
\resp PDM-Closed by $\SI{\InterPlanImprovementOverPDMClosed}{\percent}$.
This out-of-distribution gain is the clearest evidence for the arbitration graph approach:
both planners are specialized to nuPlan,
yet their composition yields its \emph{largest} improvement precisely where neither is individually reliable.
The sharpest case is the \emph{avoid crashed cars} scenario type (\cref{tab:interplan_per_scenario}),
where both constituent planners score zero on their own, yet \textit{Mosaic} reaches~$\AvoidCrashedCarsMosaicCLSR$.

These results demonstrate that the proposed AG framework is able to compose heterogeneous motion planners as an alternative to ad hoc rule-based heuristics, while preserving explainability.

\begin{table}
    \vspace*{2mm}
    \caption{Per-scenario-type CLS-R scores on Val14. Best result per row in \textbf{bold}, second-best \underline{underlined}. Rows in order of ascending \textit{Mosaic} score.}
    \centering
    \footnotesize
    \begin{tabular}{l ccc}
        \toprule
        Scenario type ($n$) & \texttt{PDM-Cl.} & \texttt{FlowDrive*} & Mosaic \\
        \midrule
Near mult.\ vehicles (85) & 89.78 & \underline{89.97} & \textbf{90.63} \\
Start.\ left turn (100) & \underline{89.42} & 89.20 & \textbf{90.78} \\
High lat.\ accel. (96) & 87.96 & \underline{91.44} & \textbf{92.28} \\
Follow.\ w/ lead (15) & \underline{94.52} & \textbf{95.10} & 92.74 \\
Changing lane (70) & \textbf{94.33} & 91.27 & \underline{92.79} \\
Trav.\ pickup/dropoff (99) & 91.59 & \underline{91.92} & \textbf{93.12} \\
Start.\ right turn (98) & 89.07 & \textbf{94.49} & \underline{93.58} \\
Low mag.\ speed (100) & 90.75 & \underline{91.86} & \textbf{94.31} \\
Start.\ straight TL int. (98) & 92.28 & \underline{94.23} & \textbf{94.34} \\
Wait.\ for pedestrian (53) & 92.71 & \underline{92.76} & \textbf{95.94} \\
High mag.\ speed (99) & 94.51 & \textbf{96.60} & \underline{96.27} \\
Stationary in traffic (98) & \underline{96.36} & 96.34 & \textbf{96.77} \\
Stopping w/ lead (93) & \textbf{98.96} & 98.24 & \textbf{98.96} \\
Behind long vehicle (14) & \textbf{100.00} & 99.11 & \textbf{100.00} \\
\midrule
Overall (1118) & 92.37 & \underline{93.37} & \textbf{94.18} \\
\bottomrule
     \end{tabular}
    \label{tab:per_scenario}
\end{table}

\begin{table}
    \vspace*{2mm}
    \caption{
        Per-scenario-type CLS-R scores on interPlan.
        Best result per row in \textbf{bold}, second-best \underline{underlined}.
        Rows sorted by \textit{Mosaic} score.
        In \textit{avoid crashed cars}, both constituent planners score zero yet \textit{Mosaic} reaches~$\AvoidCrashedCarsMosaicCLSR$.
    }
    \centering
    \footnotesize
    \begin{tabular}{l ccc}
        \toprule
        Scenario type ($n$) & \texttt{PDM-Cl.} & \texttt{FlowDrive*} & Mosaic \\
        \midrule
Constr.\ zone (10) & \textbf{18.12} & \underline{17.85} & \underline{17.85} \\
Overtake parked veh. (10) & 9.19 & \underline{26.63} & \textbf{26.85} \\
Avoid crashed cars (10) & \underline{0.00} & \underline{0.00} & \textbf{29.80} \\
Med.\ traffic density (10) & \underline{61.20} & 41.00 & \textbf{62.01} \\
High traffic density (10) & \underline{61.72} & 38.62 & \textbf{63.02} \\
Low traffic density (10) & 62.26 & \textbf{68.02} & \underline{65.08} \\
Nudge parked veh. (10) & 73.76 & \textbf{82.40} & \underline{81.51} \\
Jaywalk.\ pedestrian (10) & 53.63 & \underline{78.42} & \textbf{86.16} \\
\midrule
Overall (80) & 42.68 & \underline{44.05} & \textbf{54.10} \\
\bottomrule
     \end{tabular}
    \label{tab:interplan_per_scenario}
\end{table}

\subsection{Computational Overhead}
\label{subsec:runtime}
Averaged over the full Val14 benchmark on the same hardware,
\textit{Mosaic} spends \SI{\RuntimeMosaicPerScenarioSec}{\second} of compute per scenario,
compared to \SI{\RuntimeFlowDrivePerScenarioSec}{\second} for \texttt{FlowDrive*} alone
and \SI{\RuntimePDMClosedPerScenarioSec}{\second} for \texttt{PDM-Closed} alone.
The cost of adding the second planner is bounded because \texttt{FlowDrive*} runs on the GPU
whereas \texttt{PDM-Closed} and all arbitration logic run on the CPU, so the two planners occupy independent compute resources.
In our benchmark they are evaluated sequentially.
Because they are independent they \emph{could} be executed in parallel, though we have not measured a parallel implementation.
Verification and emergency-stop logic were already present inside each individual planner
and are merely consolidated into a shared module here,
so the only additional computation is the lightweight trajectory scoring step.

\subsection{Ablation Studies}
\label{subsec:ablations}

To isolate the contribution of each component, we conduct three ablation experiments on the CLS-R benchmark: (i) removing the verification module while keeping both planners (\textit{w/o verif.}), (ii) using only \texttt{FlowDrive*} with the shared verifier and emergency stop, and (iii) using only \texttt{PDM-Closed} with the shared verifier and emergency stop. 
Across every \textit{Mosaic} configuration the planners' native internal safety checks are disabled: the two raw-baseline rows are the only ones that retain them, the single-planner and full rows use the shared verifier instead, and \textit{w/o verif.} uses none.
These experiments are discussed in the following.

\begin{table}
    \vspace*{2mm}
    \caption{Ablation on the Val14 CLS-R benchmark. The first two rows are the \emph{original} standalone implementations of PDM-Closed and FlowDrive* with their internal verifier (speed gate at \SI{5}{\metre\per\second}, \SI{4.05}{\metre\per\square\second} emergency brake). The remaining rows share the centralized \textit{Mosaic} verifier (all speeds, \SI{\EmergencyDeceleration}{\metre\per\square\second}). \textit{Coll.}: at-fault collision count. \textit{Zero}: zero-score scenarios. \textit{EB}: \% of time steps triggering an emergency brake. Best per column in \textbf{bold}, second-best \underline{underlined}.}
    \centering
    \footnotesize
    \begin{tabular}{l|ccc|c}
        \toprule
        \multirow{2}{*}{Configuration} & \multicolumn{3}{c|}{Per Scenario} & Per Step \\
         & CLS-R $\uparrow$ & Coll. $\downarrow$ & Zero $\downarrow$ & EB $\downarrow$ \\
        \midrule
        PDM-Closed (original) & \RawPDMClosedCLSR & \RawPDMClosedColl & \RawPDMClosedZero & \RawPDMClosedEB\% \\
        FlowDrive* (original) & \RawFlowDriveCLSR & \RawFlowDriveColl & \RawFlowDriveZero & \RawFlowDriveEB\% \\
        \midrule
        Mosaic w/o verif. & \AblationWithoutVerifCLSR & \AblationWithoutVerifColl & \AblationWithoutVerifZero & - \\
        \texttt{PDM-Closed} only & \AblationPDMClosedOnlyCLSR & \AblationPDMClosedOnlyColl & \AblationPDMClosedOnlyZero & \underline{\AblationPDMClosedOnlyEB\%} \\
        \texttt{FlowDrive*} only & \underline{\AblationFlowDriveOnlyCLSR} & \textbf{\AblationFlowDriveOnlyColl} & \underline{\AblationFlowDriveOnlyZero} & \AblationFlowDriveOnlyEB\% \\
        Mosaic (full) & \textbf{\MosaicValCLSR} & \underline{\MosaicCollisions} & \textbf{\AblationMosaicZero} & \textbf{\VerifFailBothPercent\%} \\
        \bottomrule
    \end{tabular}
    \label{tab:ablation_results}
\end{table}

\begin{figure}[t]
    \centering
    \setlength{\tabcolsep}{1pt}
    \begin{tabular}{@{}c@{\hspace{1pt}}cc@{}}
         & \footnotesize $t_1$ & \footnotesize $t_2$ \\
        \adjustbox{valign=m}{\rotatebox[origin=c]{90}{\footnotesize Full \textit{Mosaic}}} &
        \adjustbox{valign=m}{\includegraphics[width=0.44\linewidth]{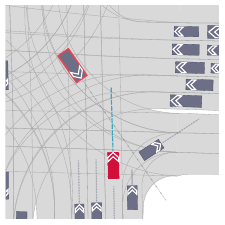}} &
        \adjustbox{valign=m}{\includegraphics[width=0.44\linewidth]{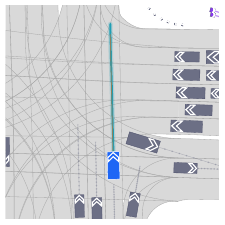}} \\
        \adjustbox{valign=m}{\rotatebox[origin=c]{90}{\footnotesize No verifier}} &
        \adjustbox{valign=m}{\includegraphics[width=0.44\linewidth]{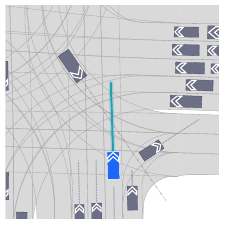}} &
        \adjustbox{valign=m}{\includegraphics[width=0.44\linewidth]{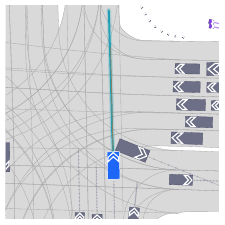}} \\
    \end{tabular}
    \caption{
        Verifier benefit in a real intersection scenario, where a late-crossing vehicle turns left across the ego's path.
        \textbf{Top:} with the shared verifier active, both planner proposals are rejected at $t_1$ (dashed)
        because each predicts contact with the crossing vehicle (outlined red).
        \textit{Mosaic} falls back to \texttt{Emergency Stop} (red) and the ego passes safely behind at $t_2$.
        \textbf{Bottom:} without verification, both proposals proceed and the ego collides at $t_2$.
        Ego in blue (red when emergency stop is active).
        The two proposals nearly coincide in this scene causing \texttt{FlowDrive*} (teal) to be drawn over \texttt{PDM-Closed} (orange).
        Dotted lines are agent predictions.
    }
    \label{fig:bev_verification}
\end{figure}

\subsubsection{Verification is the safety mechanism}
Removing verification while keeping both planners raises at-fault collisions
from \MosaicCollisions{} to \AblationWithoutVerifColl{} (\cref{tab:ablation_results}),
a \SI{\VerifCollisionReductionPercent}{\percent} reduction attributable to verification alone.
\cref{fig:bev_verification} makes the mechanism concrete:
at an intersection where a vehicle turns across the ego's path,
the verifier rejects both proposals and the ego brakes,
whereas without verification it collides.
The raw-baseline rows further show that the shared verifier is \emph{stricter} than the planners' original internal checks:
under their original configuration both \texttt{PDM-Closed} and \texttt{FlowDrive*} incur \RawPDMClosedColl{} at-fault collisions,
which the shared verifier reduces to \AblationPDMClosedOnlyColl{} and \AblationFlowDriveOnlyColl{} respectively.
Because it checks at all speeds and brakes at \SI{\EmergencyDeceleration}{\metre\per\square\second},
this stronger verification stack accounts for the safety gains.

\subsubsection{Composition is the performance mechanism}
With verification in place, composition adds no further collision reduction:
the \AblationFlowDriveOnlyColl{}, \AblationPDMClosedOnlyColl{},
and \MosaicCollisions{} at-fault collisions of \texttt{FlowDrive*}-only, \texttt{PDM-Closed}-only,
and full \textit{Mosaic} all sit at the same verification floor,
the small spread being noise at this low count.
The value of composing planners is instead in performance:
\textit{Mosaic} reaches \MosaicValCLSR{} CLS-R, above both single-planner ablations,
and reduces zero-score scenarios to \AblationMosaicZero{},
below the \AblationFlowDriveOnlyZero{} and \AblationPDMClosedOnlyZero{} of either planner alone
and the fewest of any configuration.
The two mechanisms are therefore non-redundant---a safety floor and a performance ceiling.

\subsubsection{Residual collisions}
We inspected all \MosaicCollisions{} at-fault collisions that remain in full \textit{Mosaic}
and none are caused by the arbitration logic.
Ten are benchmark artifacts unavoidable by any planner:
five scenarios initialize the ego already in or immediately before an unavoidable collision,
and five involve ghost or unstable detections that nuPlan replays as ground truth.
Four reflect limits of the constant-velocity forecast and perception noise shared by every configuration using the same verifier.
The remaining two are nuPlan attribution artifacts in which the other agent causes the collision but the geometric heuristic still flags the ego.

\subsubsection{Planner complementarity}
\begin{figure}
    \centering
    \includegraphics[width=\linewidth]{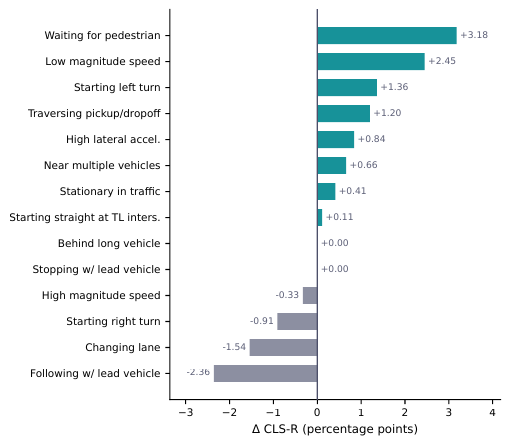}
    \caption{
        Per-scenario-type CLS-R gain of \textit{Mosaic} over the better single-planner ablation on Val14.
        Positive values mark scenario types where per-step composition outperforms either planner alone.
    }
    \label{fig:complementarity}
\end{figure}

As shown in \cref{tab:per_scenario} and \cref{fig:complementarity}, the two planners exhibit complementary strengths:
each one leads in roughly half of the 14 scenario types, with no single planner dominating across the board.
The cost arbitrator exploits this:
\textit{Mosaic} outperforms both single-planner ablations in \ScenarioTypesOutperform{} of the 14 scenario types
and matches the stronger one in \ScenarioTypesTied{} more.
Composition is not free:
in the remaining \ScenarioTypesUnderperform{} types \textit{Mosaic} falls below the better constituent,
by at most \num{\LargestScenarioTypeRegression} CLS-R points (\cref{fig:complementarity}).
These are types where one planner is consistently stronger,
so the arbitrator's occasional per-step switch to the weaker proposal pulls the average down.
The trade-off is favorable overall: \textit{Mosaic} still leads both ablations.

The same complementarity is visible at the level of individual time steps.
Each planner is decisive in a substantial share of steps.
The arbitrator selects \texttt{PDM-Closed} as the higher-scoring behavior in \SI{\SelectionPDMClosedPercent}{\percent}
and \texttt{FlowDrive*} in \SI{\SelectionFlowDrivePercent}{\percent}, so both contribute materially to the composed policy.
In the remaining \SI{\SelectionEqualPercent}{\percent} the two proposals score equally and the learning-based planner is chosen by design.
Wherever the planners disagree the arbitrator can take the stronger proposal,
which is where the per-scenario-type gains in \cref{fig:complementarity} originate.

\subsubsection{Behavior Verification}
As described in \cref{subsec:methodolgy/verification}, a trajectory is rejected when an at-fault collision is predicted.
If neither \texttt{FlowDrive*} nor \texttt{PDM-Closed} produces a verified trajectory, the \texttt{Emergency Stop} behavior is executed.
As shown in \cref{tab:ablation_results}, \texttt{PDM-Closed} fails verification in \SI{\VerifFailPDMPercent}{\percent} and \texttt{FlowDrive*} in \SI{\VerifFailFlowDrivePercent}{\percent} of time steps.
At least one behavior fails verification in \SI{\VerifFailAtLeastOnePercent}{\percent} of time steps, while both fail simultaneously in just \SI{\VerifFailBothPercent}{\percent} (\EmergencyStopCount{} out of \num{\TotalTimesteps} time steps).
The low frequency of emergency stops illustrates the key benefit of the AG: in most cases where one planner proposes an unsafe trajectory, the other provides a valid alternative, preventing unnecessary interventions.

\subsubsection{Parameter Sensitivity}
\begin{figure}
    \centering
    \includegraphics[width=0.5\linewidth]{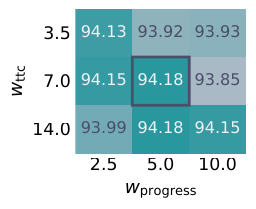}
    \caption{
        CLS-R sensitivity to $w_\mathrm{ttc}$ and $w_\mathrm{progress}$ on Val14.
        The outlined cell marks the weights used across this work.
    }
    \label{fig:sensitivity}
\end{figure}

Only the weight ratios affect the cost arbitrator's ranking,
leaving two free parameters once $w_\mathrm{comfortable}$ is fixed.
Halving and doubling $w_\mathrm{ttc}$ and $w_\mathrm{progress}$ around their defaults (\cref{tab:parameters})
gives the CLS-R grid in \cref{fig:sensitivity}:
scores vary by at most \SI{\SensitivityCLSRSpread}{} points (\SI{\SensitivityCLSRMin}{}--\SI{\SensitivityCLSRMax}{}),
with the default weights scoring highest, so the reported gains do not depend on tuning these weights to Val14.

\section{Conclusions}
\label{sec:conclusion}
We presented \textit{Mosaic},
a modular framework that fuses heterogeneous motion planners into an explainable decision structure
with built-in verification and fallback mechanisms.
By combining the rule-based PDM-Closed and the learning-based \mbox{FlowDrive*} planner,
\textit{Mosaic} sets a new state of the art on the nuPlan Val14 closed-loop benchmarks
with $\MosaicValCLSNR$ CLS-NR and $\MosaicValCLSR$ CLS-R.
On the highly interactive, out-of-distribution interPlan benchmark,
it improves by \SI{\InterPlanImprovementOverFlowDrive}{\percent} over its best constituent planner.
All this is achieved without retraining either planner or requiring additional data.
Our ablation study disentangles two independent contributions.
Centralized verification provides the safety floor,
reducing at-fault collisions from \RawPDMClosedColl{} for each standalone planner to \MosaicCollisions{}
and making every rejection decision traceable.
Per-step selection provides the performance gains and the lowest zero-score count of any configuration,
leveraging the planners' complementary strengths across scenario types.
Because the framework treats planners as black-box behavior components behind a shared interface,
it is designed to accommodate future planners and richer fallback hierarchies.
Demonstrating extensibility beyond two planners is left to future work.
 
\balance %
\printbibliography

\end{document}